%% file: main.tex
\documentclass{article}

\usepackage[dblblindworkshop, final]{neurips_2026}
\workshoptitle{Foundation Models for the Brain and Body}

\usepackage[utf8]{inputenc}
\usepackage[T1]{fontenc}
\usepackage{hyperref}
\usepackage{url}
\usepackage{booktabs}
\usepackage{amsfonts}
\usepackage{amsmath}
\usepackage{nicefrac}
\usepackage{microtype}
\usepackage{graphicx}
\usepackage{xcolor}
\usepackage{float}
\usepackage[most]{tcolorbox}

\title{The Platonic Brain:\\Representational Convergence in EEG Foundation Models}

\author{%
  Nicolas Waehner\textsuperscript{1},
  Enzo Tagliazucchi\textsuperscript{1},
  Jacobo Sitt\textsuperscript{2} \\
  \bfseries Demian Wassermann\textsuperscript{3},
  \bfseries Giovanni Franco Gabriel Marraffini\textsuperscript{3,4,*},
  \bfseries Trinidad Borrell\textsuperscript{2,*} \\[0.8em]
  \normalfont
  \textsuperscript{1}University of Buenos Aires, Buenos Aires, Argentina \\
  \textsuperscript{2}Paris Brain Institute (ICM), Sorbonne Université, INSERM, CNRS, \\
  Pitié-Salpêtrière Hospital, Paris, France \\
  \textsuperscript{3}Université Paris-Saclay, Inria, CEA, Palaiseau, France \\
  \textsuperscript{4}Sigma Nova, Paris, France \\[0.8em]
  \textsuperscript{*}These authors jointly supervised this work \\[0.3em]
  \texttt{nicolaswaehner@gmail.com}
}
\begin{document}
\maketitle
\begin{abstract}
The Platonic Representation Hypothesis (PRH) predicts that artificial neural networks converge on shared representations as scale and performance increase. We test whether these predictions hold for representations of the human brain with EEG foundation models.  Using EEG recordings from six participants watching a television series, together with aligned visual and language content, we tested whether EEG foundation-model representations converge within and across modalities as their embeddings become better at decoding an informative set of EEG markers. We found that EEG models with greater decodability showed stronger cross-family alignment in both local neighbourhood similarity and global structure, not consistently increasing with model parameter count. Cross-modal convergence was modality and metric dependent: greater EEG decodability was associated with stronger global alignment to language, but not consistently to vision models, while local alignment showed no consistent association. This dissociation may reflect local sensitivity to low-level sensory features, whereas global geometry may capture weaker shared semantic structure.

%This dissociation may reflect the sensitivity of local EEG neighbourhoods to low-level sensory features, whereas global geometry may capture weaker similarities arising from shared semantic content.

%Using EEG data paired with visual and language content of a television series watched by six subjects, we ask whether EEG foundation models converge on one another as they improve (intramodal convergence), and whether this convergence extends to vision and language models (crossmodal convergence), measuring both local neighbourhood structure and global geometry. We rank them by how well their embeddings predict sixteen established markers of cortical state. We find that EEG foundation models converge on one another both locally and globally as they get better, within and across families, and that this trend tracks representational quality rather than parameter count. However, when compared to vision and language, EEG improvement aligns only at the level of global geometry. This suggests that EEG foundation models converge toward a shared representation of brain activity that remains distinct from the \textcolor{red}{local} semantic spaces of vision and language models.

\end{abstract}

\section{Introduction}\label{sec:intro}

The Platonic Representation Hypothesis (PRH) proposes that artificial neural networks develop increasingly convergent latent representations as model scale and performance increase \citep{huh2024platonic}. This convergence has been interpreted as evidence that sufficiently trained models are capable of recovering the shared statistical structure of the world. Several mechanisms have been proposed to explain this effect, including task generality, simplicity bias, and model capacity. As models are trained on increasingly diverse tasks and data, the set of representations capable of satisfying all relevant constraints may become progressively smaller. 

%At the same time, larger models may be better able to reach this shared solution space, rather than settling on architecture-specific or suboptimal representations.

More recently, the PRH has been refined by the Aristotelian Representation Hypothesis (ARH), which argues that increasing architecture depth or width can systematically inflate representational similarity scores \citep{groger2026aristotelian}. The ARH therefore introduces a null-calibration that corrects representational similarity metrics for scale-induced inflation. Applying this calibration, it was shown that global convergence in overall geometry largely disappears, while local agreement on nearest neighbors persists. Empirical support for both the PRH and ARH has mainly come from vision and language models, although their predictions have recently begun to be examined in other data modalities \citep{borrell2026universe,yashwante2026timeseries}.

A further prediction of the PRH is that increasingly capable artificial models should become more aligned with neural representations in the human brain. Human information processing is thought to be hierarchically organized, with low-level sensory features progressively transformed into more abstract semantic representations, some of which generalize across sensory modalities \citep{dicarlo2012how, grillspector2004human, patterson2007where, lambonralph2017neural}. These semantic representations have also been linked to the distributional structure of language: concepts that occur in similar linguistic contexts tend to elicit similar neural representations \citep{harris1954distributional, mitchell2008predicting}. Thus, insofar as convergence among artificial neural networks reflects common statistical structure in the world, their representations may also converge toward the transmodal semantic representations encoded by the human brain.

This prediction can be tested by comparing representations learned by artificial neural networks with those measured from human neuroimaging data during sensory processing. Electroencephalography (EEG) is particularly well suited to investigating the temporal correspondence between neural and artificial representations because it offers high temporal resolution, is cost-effective and supports large-scale data acquisition. However, interpreting increasing artificial–neural alignment as evidence for representational convergence requires first establishing whether independently trained models of EEG itself become increasingly aligned as their scale and performance increase. Whether such intramodal convergence occurs for learned EEG representations remains an open question.

Here, we investigate whether the predictions of the PRH extend to foundation models trained directly on EEG recordings, which can now be pretrained on tens of thousands of hours of brain activity spanning multiple subjects and recording montages \citep{tegon2025femba,doner2025luna,jiang2025neurolm,elouahidi2025reve,yang2026steegformer}. Specifically, we ask whether EEG foundation models converge on similar representations as their performance improves (intramodal convergence), and whether this convergence extends beyond EEG to representations learned by vision and language models (crossmodal convergence).

\section{Setup}\label{sec:setup}

\paragraph{Data.} We use CineBrain \citep{gao2025cinebrain}, which includes EEG and fMRI from six subjects (four women, two men; aged 21–26) watching the same 180 minutes of the television series The Big Bang Theory. The EEG data was divided into non-overlapping windows of length determined by the epochs used to train each foundation model. Each window is paired with the visual stimulus the subjects saw and with the CineBrain scene captions covering it. Preprocessing details for both modalities are in App.~\ref{app:preprocessing}.

\paragraph{Models.} For EEG we use 14 models from five families: FEMBA \citep{tegon2025femba}, LUNA \citep{doner2025luna}, NeuroLM \citep{jiang2025neurolm}, ST-EEGFormer \citep{yang2026steegformer} and REVE \citep{elouahidi2025reve}. For vision we use DINOv2 \citep{oquab2023dinov2}, V-JEPA~2 \citep{assran2025vjepa2}, and VideoMAE together with its Kinetics-finetuned version \citep{tong2022videomae}. For language we use BLOOMZ \citep{bloomz} and OpenLLaMA \citep{geng2023openllama}, applied to the captions of each window. All models are frozen. Embedding extraction is detailed in App.~\ref{app:preprocessing}.

\paragraph{Metrics.} Two representations can be similar in two ways: they may share local neighbourhood structure or exhibit similar global pairwise-similarity structure. Following \citet{huh2024platonic} and \citet{groger2026aristotelian}, we use mutual $k$-nearest-neighbour overlap (mKNN) at $k=5$ for the local question and linear CKA \citep{kornblith2019cka} for the global one. mKNN finds the $k$ nearest neighbours of each time window in each of the two embedding spaces and reports the average fraction of neighbours they share (0: no shared neighbourhoods; 1: identical neighbourhoods). Linear CKA compares the pairwise-similarity structure of the two centred embeddings matrixes over all windows. It ranges from 0 to 1, and is high when two models agree on which windows are similar or dissimilar across the whole dataset. For each pair of models we report the best score over layer pairs.

\paragraph{Metrics calibration.}  
To calibrate the similarity metrics, we introduce a null model taking into account the temporal smoothness in the data. The i.i.d. permutation of \citet{groger2026aristotelian} would destroy that time correlation, drawing the null from a distribution with strictly less structure than the data, so it understates the baseline and overstates significance. Their framework is stated over an arbitrary permutation group, so we keep it and use the cyclic group of circular shifts, rigidly rotating one signal against the other. This breaks the correspondence between the two while leaving each one's own temporal structure intact, so what survives is content-specific alignment rather than shared smoothness. We state this formally in App. \ref{app:cyclic_null}. 
\vspace{-1.5mm}

\paragraph{A task-free Quality axis.}\label{sec:score} 

Testing whether capability
predicts alignment requires a measure of model performance for a relevant task, as in \citet{huh2024platonic}
and \citet{groger2026aristotelian}. Vision and language offer both a direct measure,
such as ImageNet accuracy or bits per byte, and an indirect measure in model size,
which scaling laws tie to loss and downstream performance
\citep{kaplan2020scaling,hoffmann2022chinchilla}. EEG models have neither: they are
orders of magnitude smaller and no scaling law relates their size to performance, and
\citet{yang2026steegformer} find performance varying widely across downstream tasks. Due to this variability across tasks, we focus on the decodability of very general EEG features that are consistently informative about global brain state across cognitive tasks and populations \citep{sitt2014large,engemann2018robust}: band power, spectral entropy, median and edge frequency, permutation entropy, and Hjorth mobility and complexity. Each marker is predicted by ridge regression, cross-validated over contiguous blocks so nearby windows never fall on both sides of a split, and a model's score is its mean over markers. We define \emph{NICE decodability} as the mean cross-validated $R^2$ with which a model's embeddings predict these markers. Details in App.~\ref{app:markers}. We use NICE decodability as our measure of capability and treat parameter count separately, so that convergence with capability and convergence with scale can be assessed independently.

\vspace{-2mm}
\paragraph{Comparing within a family.} The five model families differ in architecture, pretraining data, and objective, producing family-specific baselines in both decodability and representational alignment. To isolate the within-family association, we defined each model's alignment as its mean calibrated similarity to models from other families and subtracted the corresponding family mean from both alignment and EEG decodability. We then computed Spearman's correlation across these family-centred values.

\section{Results} 

\subsection{Intramodality Alignment}\label{sec:axis1}

\begin{figure}[H]
  \centering
  \includegraphics[width=\textwidth]{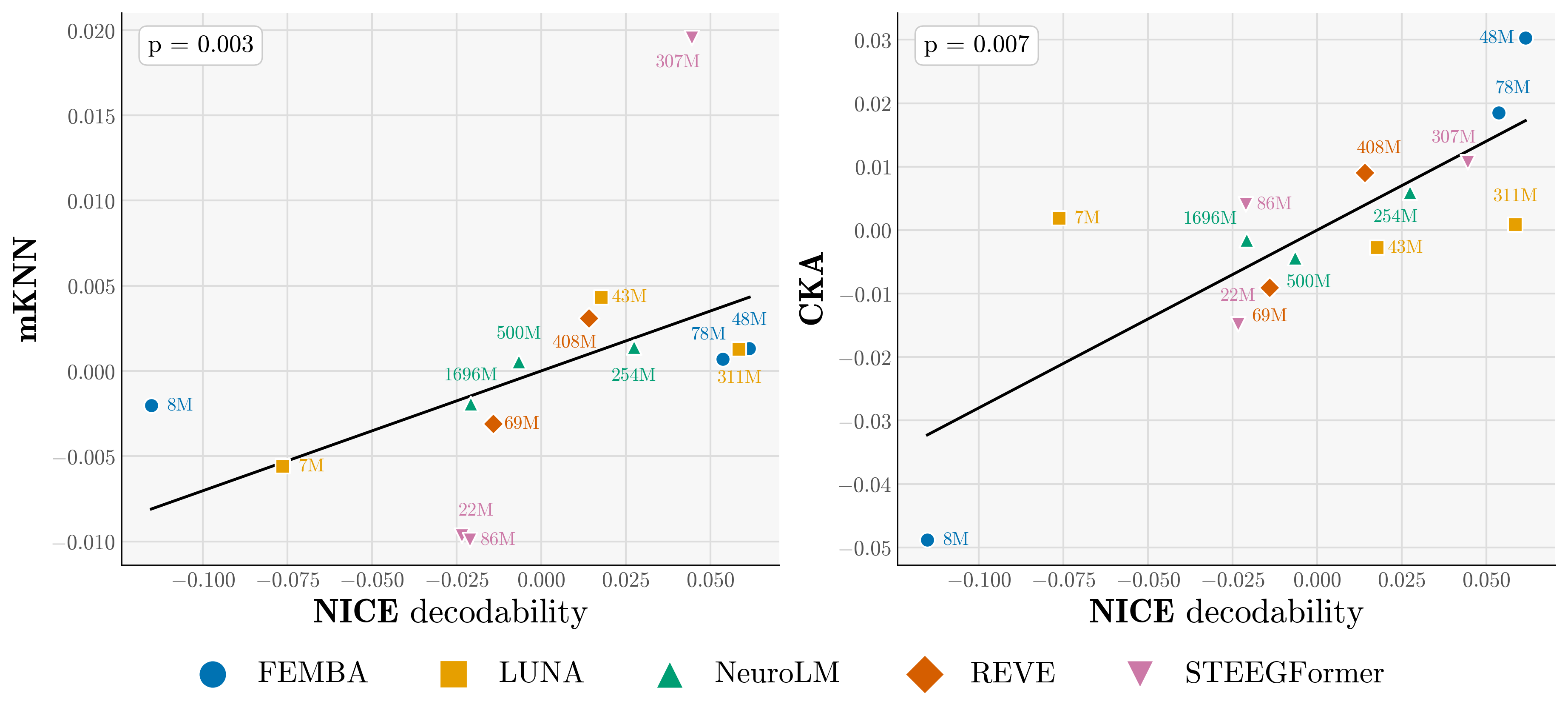}
   \vspace{-1.7em}
\caption{\textbf{Intra-modality alignment correlates with model capability.} Individual points denote models; shape and colour indicate family, and labels indicate parameter count. \emph{Vertical axis:} alignment after subtracting results obtained using the null model (App.~\ref{app:cyclic_null}), averaged over that model's alignments to every model outside its own family for the mKNN (left) and CKA (right). \emph{Horizontal axis:} NICE decodability. Both axes are mean-centred within family. Alignments are computed on subject-averaged embeddings resampled to the shared 10\,s windows ($n = 1080$).}
  \label{fig:axis2}
\end{figure}

\textbf{EEG marker decodability predicts cross-family alignment.}  As a first test of the predictions put forward by the PRH, we next tested whether models with greater NICE decodability aligned more strongly with models from other EEG families\footnote{As a preliminary analysis, we compared adjacent model-size pairs within each family (App.~\ref{app:axis1}). Calibrated mKNN increased from the smaller to the larger pair in every family, whereas CKA showed no consistent increase. Because models within a family share architectures, training data, and objectives, we excluded these comparisons from the main analysis for both metrics.}. For each model, we averaged its excess alignment over the cyclic-null mean across all cross-family comparisons and mean-centred both alignment and decodability within family (Figure~\ref{fig:axis2}). There, each panel asks: \textit{within a family, do models with greater marker decodability align more with other families?}

Decodability correlated positively with both mKNN ($p=0.003$) and CKA ($p=0.007$), consistent with the predictions of the PRH\footnote{Parameter count did not consistently order the models. For EEG models, size appears not to be a proxy for model quality.}. At the pairwise level, 78/78 mKNN and 75/78 CKA comparisons exceeded the cyclic null at $p<0.05$.

%For each model we averaged its similarity to all models outside its own family. Figure~\ref{fig:axis2} shows local (mKNN) and global (CKA) representational similarity as a function of EEG marker decodability performance. Similarity in this figure is the average excess over the cyclic null of App.~\ref{app:cyclic_null}, and both axes are mean-centred within family, so each panel asks: \textit{within a family, do better models align more with other families?} Every pairwise comparison satisfies $p < 0.05$ against the cyclic null. \footnote{Parameter count does not order the points: within every family the largest model is not the most aligned. For EEG models, size appears not to be a proxy for model quality.}

%Both local and global metrics show statistically significant correlations with NICE decodability. For each architecture, alignment between foundation models increased with their capability to decode the NICE markers, consistently with the predictions of the PRH and ARH.

\subsection{Crossmodality Alignment: Vision and Language}\label{sec:axis2}

\begin{figure}[h]
  \centering
  \includegraphics[width=\textwidth]{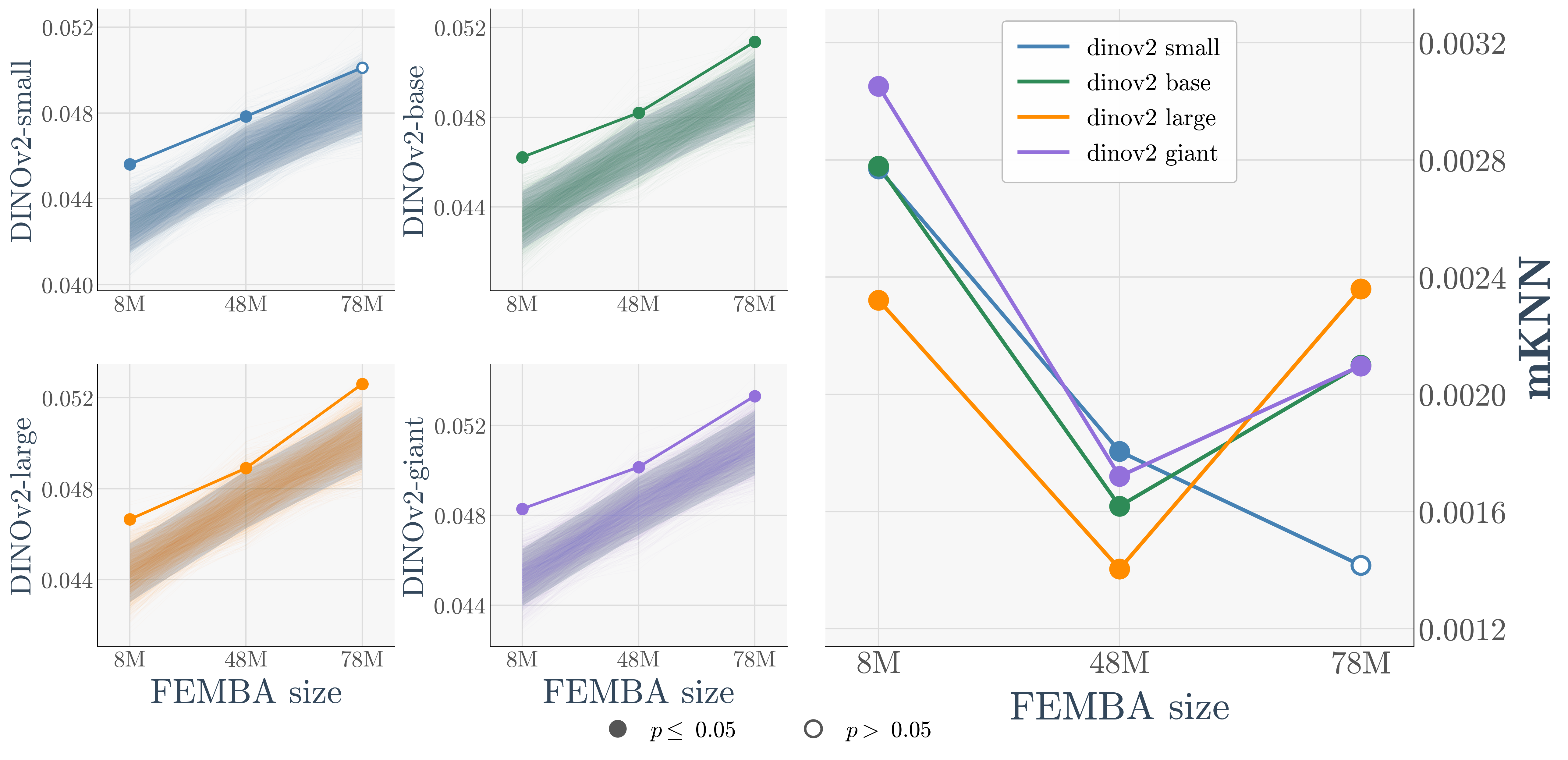}
   \vspace{-1.7em}
    \caption{\textbf{Cross-modal alignment increases due to heightened model sensitivity to temporal correlations and not because of model scaling} \emph{Left:} observed mKNN for FEMBA–DINOv2 pairs (solid coloured lines) and alignment under circular shifts (translucent lines; shaded 5th–95th percentile interval). Filled markers indicate comparisons with $p \leq 0.05$ under the cyclic-shift test. \emph{Right:} Calibrated alignment for each pair. Although raw alignment increases with FEMBA size, the null-centred value does not increase consistently. All models pairs comparison are presented in Figures \ref{fig:all_vision_lang_calibrated_mknn} and \ref{fig:all_vision_lang_calibrated_cka}.}
  \label{fig:femba_dinov2}
\end{figure}

The cyclic null introduced in Sec.~\ref{sec:setup} and applied to the within-EEG analysis in Figure~\ref{fig:axis2} was also used for all cross-modal comparisons. Figure~\ref{fig:femba_dinov2} illustrates its effect in this setting. Raw mKNN between FEMBA and DINOv2 increases with EEG model size, but the cyclic-null baseline increases at a comparable rate. In this case every cell except one clears the null model, so the alignment is real, but the excess over the null model does not increase with model size. Thus, raw alignment against model size cannot by itself establish cross-modal convergence. All model-pair comparisons are in App. \ref{app:grid_vision_net}

\begin{figure}[h]
  \centering
  \includegraphics[width=\textwidth]{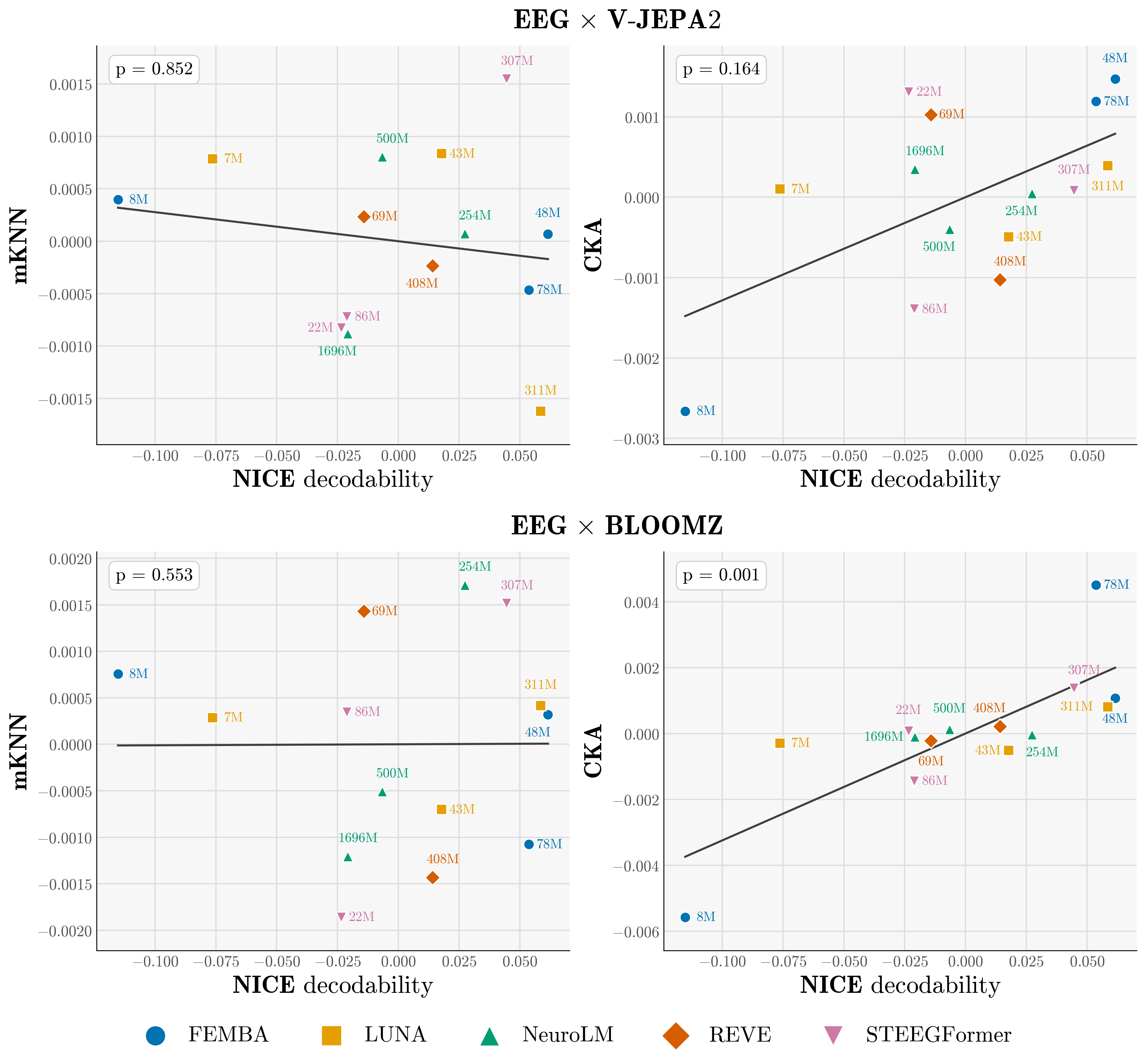}
  \vspace{-1.7em}
\caption{\textbf{Cross-modality global alignment correlates with model capability}
Individual points denote models; shape and colour indicate family, and labels indicate parameter count. \emph{Vertical axis:} alignment after subtracting results obtained using the null model (App.~\ref{app:cyclic_null}), and averaged for each EEG model across all sizes of the corresponding vision or language model family for the mKNN (left) and CKA (right). \emph{Horizontal axis:} NICE decodability. Both axes are mean-centred within family. Each EEG model is paired with the partner tiled at its own
window length on subject-averaged EEG embeddings.}
  \label{fig:vision_and_language_vs_decodability}
\end{figure}

We therefore tested cross-modal convergence using the same null-centred alignment and NICE-decoding axis as in Figure~\ref{fig:axis2}. Figure \ref{fig:vision_and_language_vs_decodability} shows alignment with V-JEPA2 and BLOOMZ, averaged across the sizes of each partner family, so the question now is: \textit{do models with greater decodability align more strongly with the vision and language representations?} Results against VideoMAE, VideoMAE-K400, DinoV2 and OpenLLaMa are presented in App \ref{app:crossmodality vis and lang}. 

Alignment does not consistently correlate with NICE decodability for either vision or language models when measured with mKNN. In contrast, CKA shows a more consistent relationship with decodability, particularly for language models. This suggests a dissociation between local and global measures of representational similarity: the relationship between alignment and decodability is apparent at the level captured by CKA, but is not consistent when measuring local alignment between models using mKNN.

%Each EEG model is paired with a vision and language model on the temporal windows defined by the EEG foundation model, scored with mKNN and CKA using the pair of layers with the highest similarity, and calibrated against the cyclic null. We plot the alignment against the same decodability axis used in \S\ref{sec:axis1}, so the question now is: \textit{do more capable models (in terms of NICE marker decodability) are more aligned with the vision and language representations?} 

%In Figure \ref{fig:vision_and_language_vs_decodability} we show the results for V-JEPA2 and BLOOMZ, averaging all different sizes within each model family. Results against VideoMAE, VideoMAE-K400, DinoV2 and OpenLLaMa are presented in App \ref{app:crossmodality vis and lang}. We observe that alignment does not consistently correlate with NICE decodability for either vision or language models when measured with mKNN. In contrast, CKA shows a more consistent relationship with decodability, particularly for language models: significant (p=0.001) for BLOOMZ and approaching statistical significance (p=0.16) for V-JEPA2. This suggests a dissociation between local and global measures of representational similarity: the relationship between alignment and decodability is apparent at the level captured by CKA, but is not consistent when measuring local alignment between models using mKNN.

\section{Discussion and limitations}

We found that representations of EEG foundation models become more similar as these models increase their capability to decode a highly informative set of EEG markers.  This was observed within and across families. We emphasise that this is evidence of convergence with capability, as measured by NICE decodability, rather than with scale. Within families, larger models showed higher calibrated local alignment (mKNN) but not global alignment (CKA), a comparison confounded by shared architectures and training data (App.~\ref{app:axis1}). Whether EEG representations converge with scale per se, as the PRH predicts for vision and language, remains open and may require EEG models spanning a wider range of sizes.

Additionally, cross-modal convergence was metric and modality dependent. Global alignment correlated strongly with decodability for language models and more weakly for some vision models, whereas local alignment showed no consistent association. One possible explanation is that low-level sensory features dominate local EEG neighbourhoods, whereas CKA summarizes global geometry across all samples and may therefore capture weaker, distributed similarities associated with shared semantic content. This interpretation is consistent with our data but was not tested directly.

If EEG foundation models converge, this has practical consequences beyond the PRH itself. First, it suggests that what these models learn reflects structure in brain activity rather than specific structure of a given architecture or pretraining pipeline, so interpretations of one model's representations (e.g., which brain states it separates) are more likely to transfer to others. Second, alignment to other models together with marker decodability offers a task-free way to compare EEG foundation models, which is useful where downstream benchmarks disagree and labelled data are scarce, as is common in clinical settings. Third, the global alignment with language models points to language as a possible common reference frame for interpreting EEG representations.

Our results and conclusions contain clear limitations. The analyses included only 14 models from five families and six participants, limiting statistical power. Also, neighbourhood similarity between EEG and video or language representations exceeded the time-shifted control only modestly, suggesting that part of the observed alignment may reflect shared temporal smoothness rather than content-specific structure. Finally, all analyses relied on a single dataset, limiting the generalisability of our conclusions.

\newpage

\begin{ack}
This work was supported by a BlueDot Rapid Grant.
Nicolas Waehner is supported by a CONICET doctoral fellowship.
Trinidad Borrell is supported by Paris Brain Institute America.
Giovanni Marraffini is employed by Sigma Nova. 

The authors declare no other competing interests.
\end{ack}

\bibliographystyle{plainnat}
\bibliography{refs}

\appendix

\input{appendix_tables}

\end{document}

%% file: appendix_tables.tex
\section{Preprocessing}
\label{app:preprocessing}

\paragraph{Data and windowing.} We use Season~7 of CineBrain
\citep{gao2025cinebrain}, the first ten episodes, which all six subjects share:
$10\,800$~s stored as 13\,500 consecutive $0.8$~s frames. The EEG is recorded
with a 64-channel GSN-HydroCel cap at $1000$~Hz and distributed band-pass
filtered to $0.1$--$30$~Hz with a $50$~Hz notch; the fMRI is distributed as one
flat masked-voxel vector of $18\,946$ dimensions per repetition time
($\mathrm{TR}=0.8$~s), already z-scored and not atlas-parcellated. Both share
the same $0.8$~s frame grid, on which a $4$~s video clip $c$ covers frames
$[5c, 5c+5)$. We take both releases as distributed and add no cleaning of our
own: no independent component analysis, no artefact rejection, no channel
interpolation, and no re-referencing beyond the montages individual models
require, so every window of every subject enters the analysis. All
representations live on non-overlapping windows tiling the recording from $t=0$,
with the length $L$ fixed by a window consistent with the ones used in pretraining for each EEG model family:
$5$~s for FEMBA and LUNA ($W=2160$ windows), $6$~s for ST-EEGFormer ($W=1800$),
$8$~s for NeuroLM ($W=1350$), $10$~s for REVE ($W=1080$), and a $4$~s
clip-native grid ($W=2700$) where no EEG model is involved.

\paragraph{EEG.} Each model receives the input format of its own pretraining
pipeline. FEMBA and LUNA use the 22 bipolar pairs of the TUEG double-banana
montage at $256$~Hz, the former with per-channel interquartile-range
normalisation and clipping to $[-20,20]$, the latter z-scored per channel and
window; NeuroLM uses 64 channels at $200$~Hz, rescaled to the microvolt range of
its training data and tokenised as 64 channels $\times$ 8 patches of $200$
samples; ST-EEGFormer uses 62 channels at $128$~Hz, z-scored per channel;
REVE uses 64 channels at $200$~Hz, band-passed $0.5$--$99$~Hz, z-scored per
channel over the full session and clipped at $\pm 15\sigma$. The models
therefore do not all see the same electrodes, and the two band-passes are inert
given the $30$~Hz low-pass of the data. For each model we extract the hidden
state of every block and mean-pool it over tokens, giving one vector per layer,
window and subject.

\paragraph{fMRI.} Treating fMRI as a representation rather than a target, it
passes through no model. Per subject we assemble the $13\,500 \times 18\,946$
matrix of per-$\mathrm{TR}$ voxel vectors and average the $\mathrm{TR}$s whose
centre time falls in $[wL + h, (w+1)L + h)$, where $h$ is a hemodynamic delay.
As the released data are already hemodynamically aligned, we swept
$h \in [0,8]$~s and found $h=0$~s to maximise alignment; all reported results use
$h=0$~s. This gives one voxel vector per window per subject, on the same grid as
every other representation.

\paragraph{Stimulus and subjects.} For a window of length $L$ we concatenate the
frames of every $4$~s clip it overlaps and sample from that span: 16 frames
uniformly for the video transformers, one frame per second for the image models,
whose per-block class tokens are averaged over the window. Language input uses
the Qwen-2.5-VL-7B caption CineBrain provides per clip; a window takes every
caption it overlaps, concatenated in order, two for the $5$, $6$ and $8$~s grids
and three for the $10$~s grid, embedded as layerwise masked mean-pools. When $L$
is not a multiple of $4$~s adjacent windows share a boundary clip and caption,
inflating temporal autocorrelation, which the shift-null and block-permutation
tests of the main text control for. 
Alignment scores are computed per subject
and then averaged, so no similarity computation mixes subjects, whereas the
interpretability analyses use subject-averaged embeddings and markers.

\section{Null calibration}\label{app:cyclic_null}

Here we define the null used in this work, adapted from the i.i.d calibration of \citet{groger2026aristotelian}.

Stating this formally: let $n$ be the number of windows in the grid, and let
$C_n = \{c_d\}_{d=0}^{n-1}$ be the cyclic group whose elements rigidly rotate the
window index,
\begin{equation}
  c_d(\mathbf{Y})[i] \;=\; \mathbf{Y}\!\left[(i-d) \bmod n\right] .
\end{equation}
That is, $c_d$ moves the entry at position $i-d$ to position $i$: the whole sequence
slides by $d$ windows, and the $\bmod n$ returns what falls off the end to the front.
The rotated sequence holds the same $n$ vectors in the same cyclic order, so
$\mathbf{Y}$'s own autocorrelation is left intact and its marginal distribution is
identical; only its correspondence with $\mathbf{X}$ is destroyed, whereas a freea
shuffle destroys both.

The cyclic group defines a strictly stronger null than the i.i.d. one:
\begin{tcolorbox}[
  colback=gray!15,
  colframe=gray!50,
  boxrule=0.5pt,
  arc=2pt
]
\noindent
\textbf{$H_0$:} $\mathbf{X}$ and $\mathbf{Y}$ similarity at the true temporal pairing is no greater than at an arbitrary circular shift of one against the other.
\end{tcolorbox}

Assumption 3.1 of \citet{groger2026aristotelian} then reads
$P_{H_0'}(\mathbf{X},\mathbf{Y}) = P_{H_0'}(\mathbf{X}, c_d(\mathbf{Y}))$ for every
$c_d \in C_n$, i.e.\ approximate circular stationarity of $\mathbf{Y}$.

Because we report the maximum alignment $s$ over layer pairs, we follow the aggregation-aware
calibration of \citet[\S5.2]{groger2026aristotelian}: the \emph{same} rotation is
applied to every layer of $B$ before the maximum is taken,
\begin{equation}
  S^{(d)}_{\ell,\ell'} \;=\; s\!\left(\mathbf{X}^{(A)}_{\ell},\,
  c_d\!\left(\mathbf{Y}^{(B)}_{\ell'}\right)\right), \qquad
  T^{(d)} \;=\; \max_{\ell,\ell'} S^{(d)}_{\ell,\ell'} ,
\end{equation}
so the null retains the same selection advantage as the observed statistic. 

Since $c_0$ is the identity, $T_{\mathrm{obs}} = T^{(0)}$, and $|C_n| = n$ is small
enough to enumerate exhaustively, so $K = n-1$ is fixed by the number of windows rather than chosen and the right-tail permutation $p$-value
\begin{equation}\label{eq: perm test}
  p \;=\; \frac{1 + \#\{d \neq 0 \,:\, T^{(d)} \ge T^{(0)}\}}{K+1}
\end{equation}
is exact rather than Monte Carlo, with floor $1/n$. 
We report as magnitude the excess over the null
\begin{equation}
  \Delta \;=\; T^{(0)} \;-\; \frac{1}{n-1}\sum_{d=1}^{n-1} T^{(d)} .
\end{equation}
in every alignment plot, whether it is mKNN or CKA. $\Delta$ is the ``null-centered'' calibration that
\citet[App.~E.3]{groger2026aristotelian} evaluate alongside their gated score and find
to collapse correctly to the null baseline across all metrics; we prefer it here because
it is $\alpha$-free and, unlike their Eq.~(17), commutes with the averaging over
comparisons that our $y$ axes perform in Figures \ref{fig:axis2}, \ref{fig:vision_and_language_vs_decodability}, \ref{fig:crossmod_lang} and \ref{fig:crossmod_vis}. Evidence against $H_0$ is carried by $p$.

\section{NICE markers}\label{app:markers}

The markers span three families: \textbf{(i) Spectral (11 markers)}: absolute
(log) and normalised band powers in $\delta$ (1--4~Hz), $\theta$ (4--8~Hz),
$\alpha$ (8--13~Hz) and $\beta$ (13--30~Hz), plus spectral entropy, median
spectral frequency and the $90\%$ spectral edge frequency (SE$_{90}$), all
computed from a Welch periodogram restricted to 1--30~Hz;
\textbf{(ii) Information theory (3 markers)}: permutation entropy
\citep{Bandt2002} of the broadband signal and of the $\theta$- and
$\alpha$-filtered signals (embedding order $3$, unit delay, normalised by
$\log 3!$); \textbf{(iii) Complexity (2 markers)}: the Hjorth mobility and
complexity parameters, i.e.\ the variance ratios of the signal and its first
and second derivatives. 

Each marker is computed per channel and then summarised as the median across
the 64 EEG channels, yielding one scalar per marker per time window. Windows
are non-overlapping and tile the recording from $t=0$; their length matches the input window of the EEG model under comparison (5, 6, 8 or 10~s), so that
marker vectors and model embeddings are defined on exactly the same temporal
grid. The resulting per-window vectors are averaged across the six subjects, to match the subject-averaged embeddings. This produces a 16-dimensional
embedding per window. Crucially, every dimension is interpretable by
construction, corresponding to a specific marker whose clinical meaning has
been established across years of research \citep{engemann2018robust}.

\section{Models, grids and marker scores}\label{app:models}

Table~\ref{tab:app-models} lists every EEG model, the grid it defines, its marker score, and its mean
similarity to models of other families. Table~\ref{tab:app-markers} breaks the marker score down by
marker.
\begin{table}[H]\centering\small
\caption{The 14 EEG models. Window and count give the grid each model defines natively; cross-model similarity is computed after resampling to the shared
10\,s grid (1080 windows). The marker score is the mean $R^2$ over the sixteen EEG markers
on subject-averaged data, and is the horizontal axis of Figures~\ref{fig:axis2}
and~\ref{fig:vision_and_language_vs_decodability}. Values are uncentred; the figures plot the same quantities after mean-centring within family.}
\label{tab:app-models}
\begin{tabular}{llrrrrrr}
\toprule
Family & Size & Params (M) & Window (s) & Windows & Marker $R^2$ & MKNN & CKA \\
\midrule
FEMBA       & tiny  &    8 &  5 & 2160 & $-0.154$ & 0.057 & 0.136 \\
FEMBA       & base  &   48 &  5 & 2160 & $+0.023$ & 0.061 & 0.215 \\
FEMBA       & large &   78 &  5 & 2160 & $+0.015$ & 0.060 & 0.203 \\
LUNA        & base  &    7 &  5 & 2160 & $+0.190$ & 0.063 & 0.284 \\
LUNA        & large &   43 &  5 & 2160 & $+0.284$ & 0.073 & 0.279 \\
LUNA        & huge  &  311 &  5 & 2160 & $+0.325$ & 0.070 & 0.283 \\
NeuroLM     & b     &  254 &  8 & 1350 & $+0.594$ & 0.041 & 0.226 \\
NeuroLM     & l     &  500 &  8 & 1350 & $+0.560$ & 0.040 & 0.215 \\
NeuroLM     & xl    & 1696 &  8 & 1350 & $+0.546$ & 0.037 & 0.218 \\
REVE        & base  &   69 & 10 & 1080 & $+0.667$ & 0.052 & 0.223 \\
REVE        & large &  408 & 10 & 1080 & $+0.695$ & 0.058 & 0.241 \\
STEEGFormer & small &   22 &  6 & 1800 & $+0.663$ & 0.049 & 0.287 \\
STEEGFormer & base  &   86 &  6 & 1800 & $+0.666$ & 0.049 & 0.306 \\
STEEGFormer & large &  307 &  6 & 1800 & $+0.731$ & 0.078 & 0.313 \\
\bottomrule\end{tabular}\end{table}

\begin{table}[H]\centering\small
\caption{The sixteen EEG markers, and how well the model embeddings predict them. Median and best
$R^2$ are taken across the 14 EEG models. The three permutation entropy markers had non-positive median ($R^2$) across models, although some individual models predicted them above chance.}
\label{tab:app-markers}
\begin{tabular}{lrrl}
\toprule
Marker & Median $R^2$ & Best $R^2$ & Best model \\
\midrule
abs\_delta & +0.466 & +0.887 & reve-base \\
rel\_delta & +0.676 & +0.768 & reve-large \\
abs\_theta & +0.232 & +0.877 & reve-large \\
rel\_theta & +0.523 & +0.631 & steegformer-large \\
abs\_alpha & +0.066 & +0.824 & reve-large \\
rel\_alpha & +0.593 & +0.671 & steegformer-large \\
abs\_beta & +0.046 & +0.850 & reve-large \\
rel\_beta & +0.345 & +0.827 & steegformer-large \\
spec\_entropy & +0.409 & +0.719 & steegformer-large \\
median\_freq & +0.597 & +0.756 & steegformer-large \\
spec\_edge90 & +0.449 & +0.786 & steegformer-large \\
pe\_broad & -0.275 & +0.779 & steegformer-large \\
pe\_theta & -0.022 & +0.198 & neurolm-xl \\
pe\_alpha & -0.025 & +0.292 & neurolm-b \\
hjorth\_mobility & +0.660 & +0.944 & steegformer-large \\
hjorth\_complexity & +0.584 & +0.868 & steegformer-large \\
\bottomrule\end{tabular}\end{table}

\section{Intramodality Intra-architecture} \label{app:axis1}

\begin{figure}[H]
  \centering
  \includegraphics[width=1\textwidth]{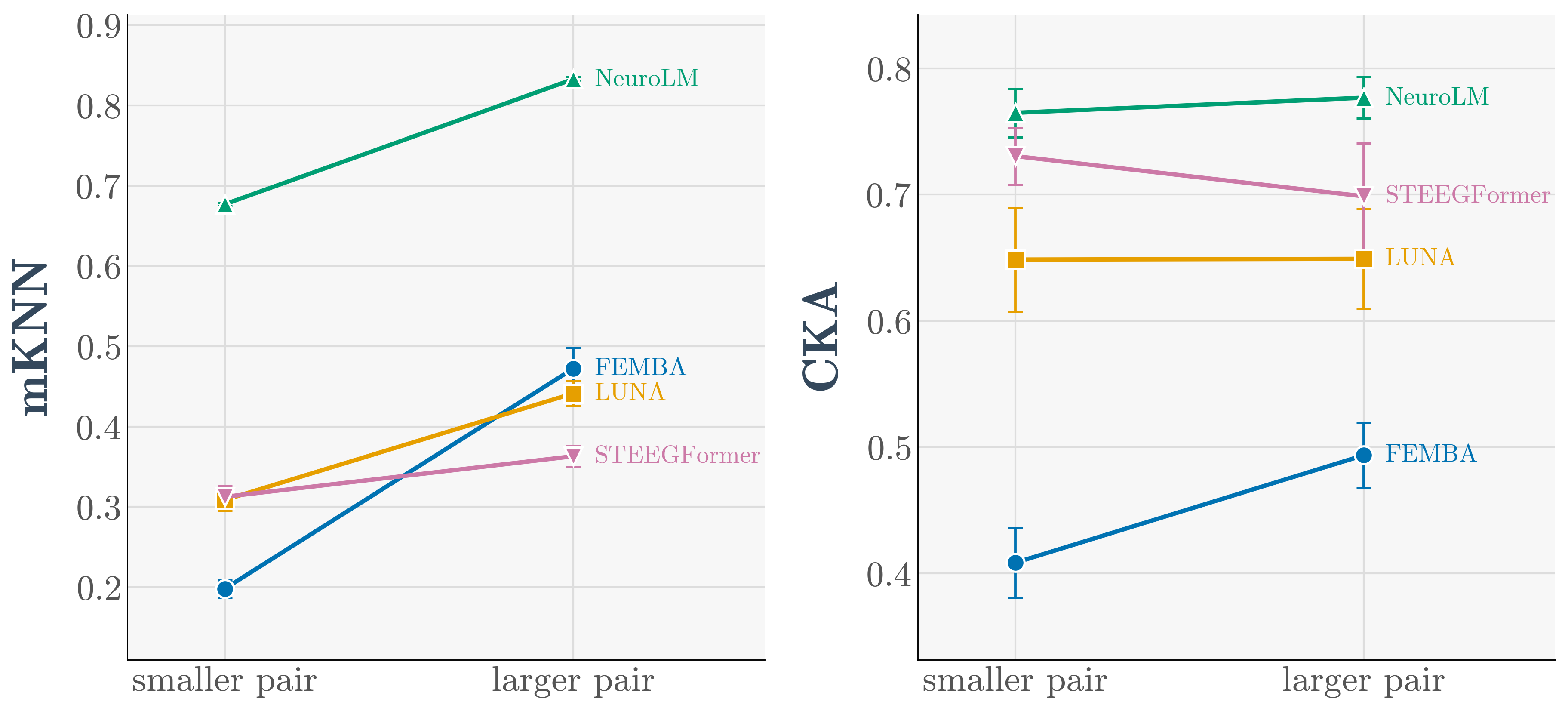}
  \caption{Similarity between two models of the same family, for the two smallest
  models and the two largest. Every family alignment rises locally but only FEMBA does globally. }
  \label{fig:axis1}
\end{figure}

For each family, we selected the two smallest models and the two largest, and measured the similarity of each pair. Figure~\ref{fig:axis1} shows how every family the larger pair is more aligned in mKNN, but only FEMBA increases in CKA. We excluded REVE because it includes only two distinct sizes.

\section{Crossmodality against every model: Vision and Language} \label{app:crossmodality vis and lang}

\begin{figure}[H]
  \centering
  \includegraphics[width=\textwidth]{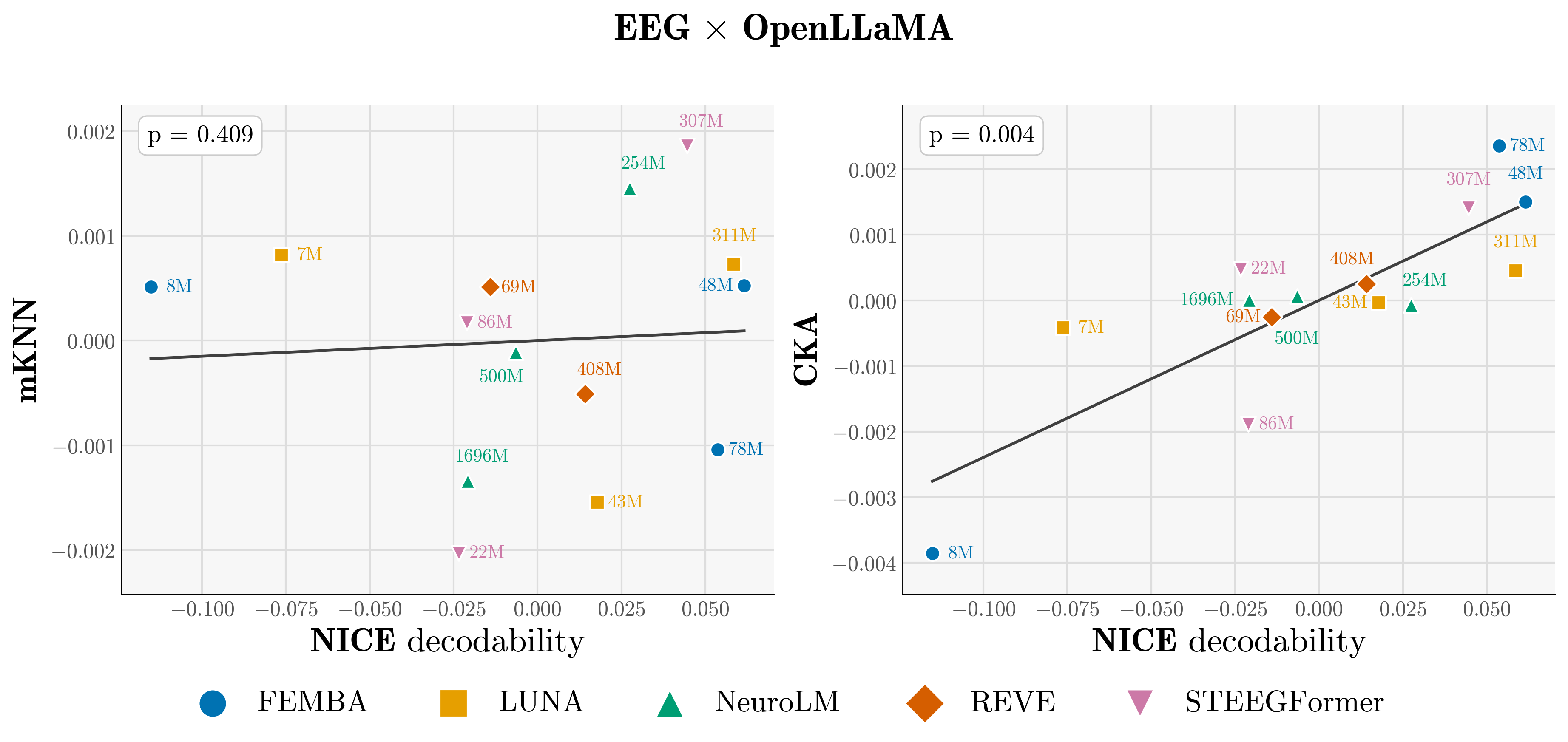}
   \vspace{-1.7em}
    \caption{Individual points denote models; shape and colour indicate family, and labels indicate parameter count. \emph{Vertical axis:} calibrated alignment averaged across the available OpenLLaMA sizes for the mKNN (left) and CKA (right). \emph{Horizontal axis:} NICE decodability. Both axes are mean-centred within family. Each EEG model is paired with the partner tiled at its own
window length ($W = 2160$, $1800$, $1350$, $1080$ for the 5, 6, 8 and 10\,s families), on subject-averaged EEG embeddings.}
  \label{fig:crossmod_lang}
\end{figure}

\begin{figure}[H]
  \centering
  \includegraphics[width=\textwidth]{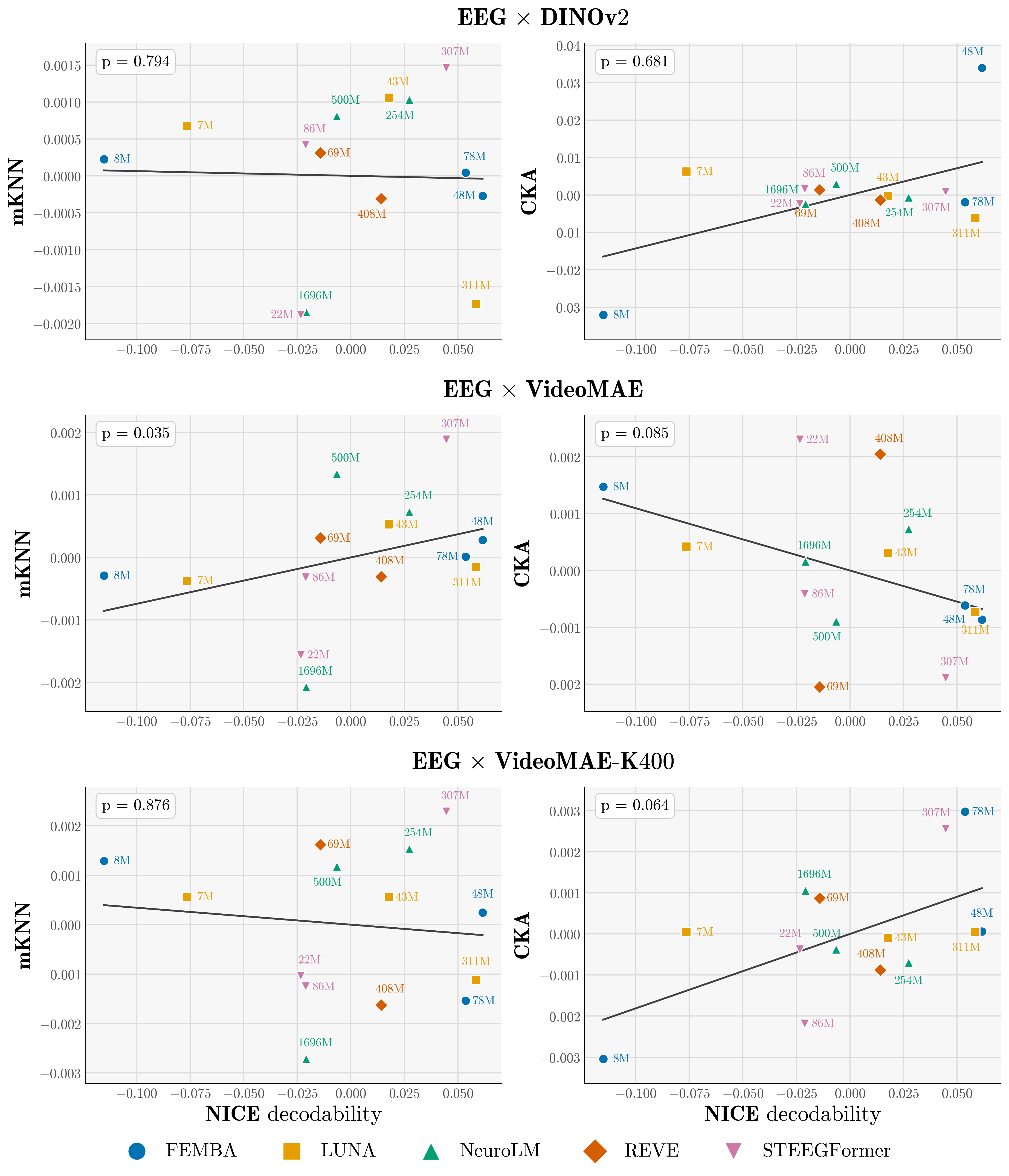}
   \vspace{-1.7em}
    \caption{Individual points denote models; shape and colour indicate family, and labels indicate parameter count. \emph{Vertical axis:} alignment calibrated averaged across all sizes of the indicated vision model family for the mKNN (left) and CKA (right). \emph{Horizontal axis:} NICE decodability. Both axes are mean-centred within family. Each EEG model is paired with the partner tiled at its own
window length ($W = 2160$, $1800$, $1350$, $1080$ for the 5, 6, 8 and 10\,s families), on subject-averaged EEG embeddings.}
  \label{fig:crossmod_vis}
\end{figure}

Here we show the results obtained with OpenLLaMA in Figure \ref{fig:crossmod_lang} and with DINOv2, VideoMAE and VideoMAE-K400 in Figure \ref{fig:crossmod_vis}, against NICE decodability. We observe consistent global alignment against language models.

\section{EEG similarity to vision and language}\label{app:grid_vision_net}

\begin{figure}[h]
  \centering
  \includegraphics[width=\textwidth]{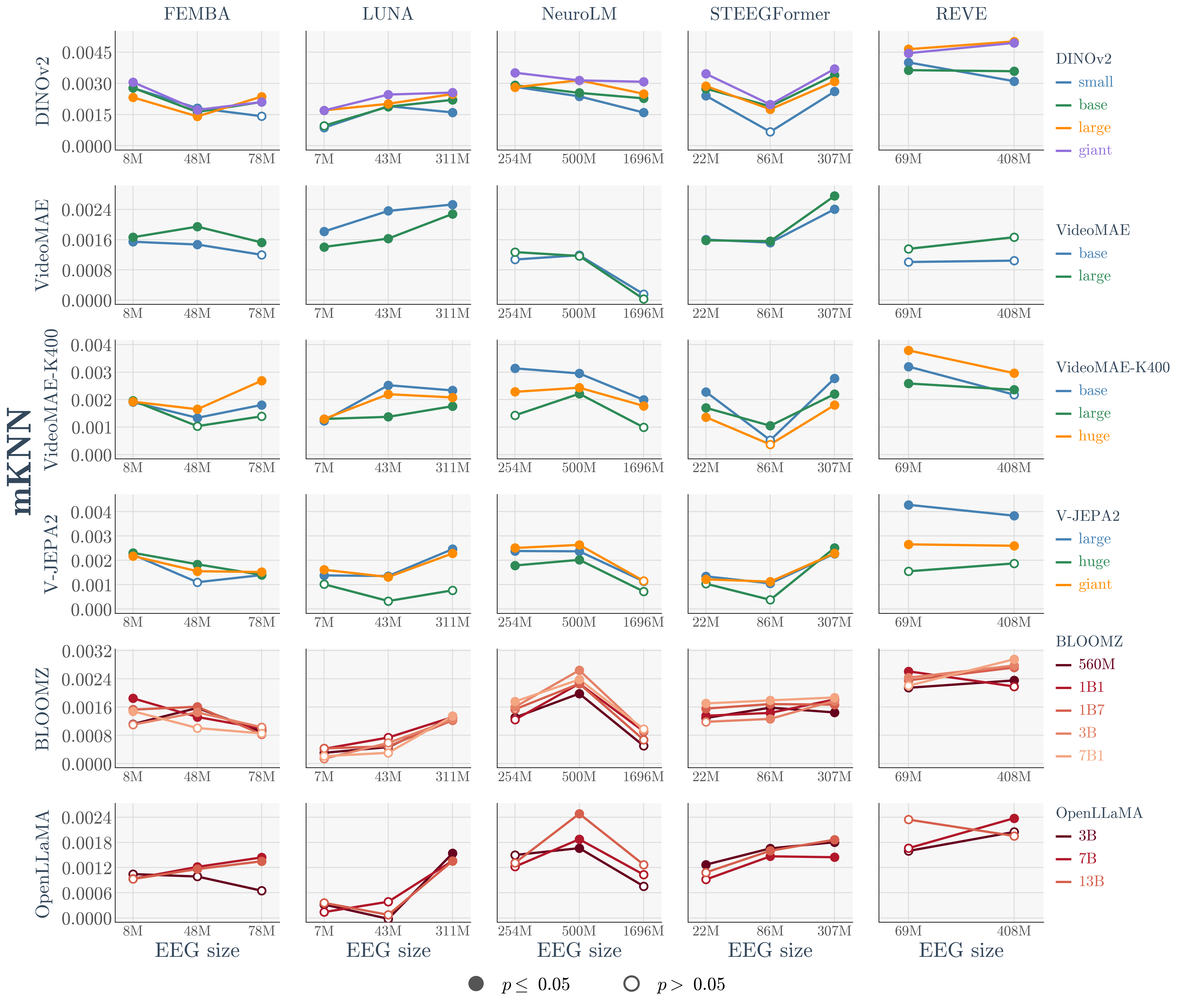}
  \caption{Each EEG model mKNN alignment to every vision and language model. Many clear the null but there is no correlation with model size.
  }
  \label{fig:all_vision_lang_calibrated_mknn}
\end{figure}

\begin{figure}[h]
  \centering
  \includegraphics[width=\textwidth]{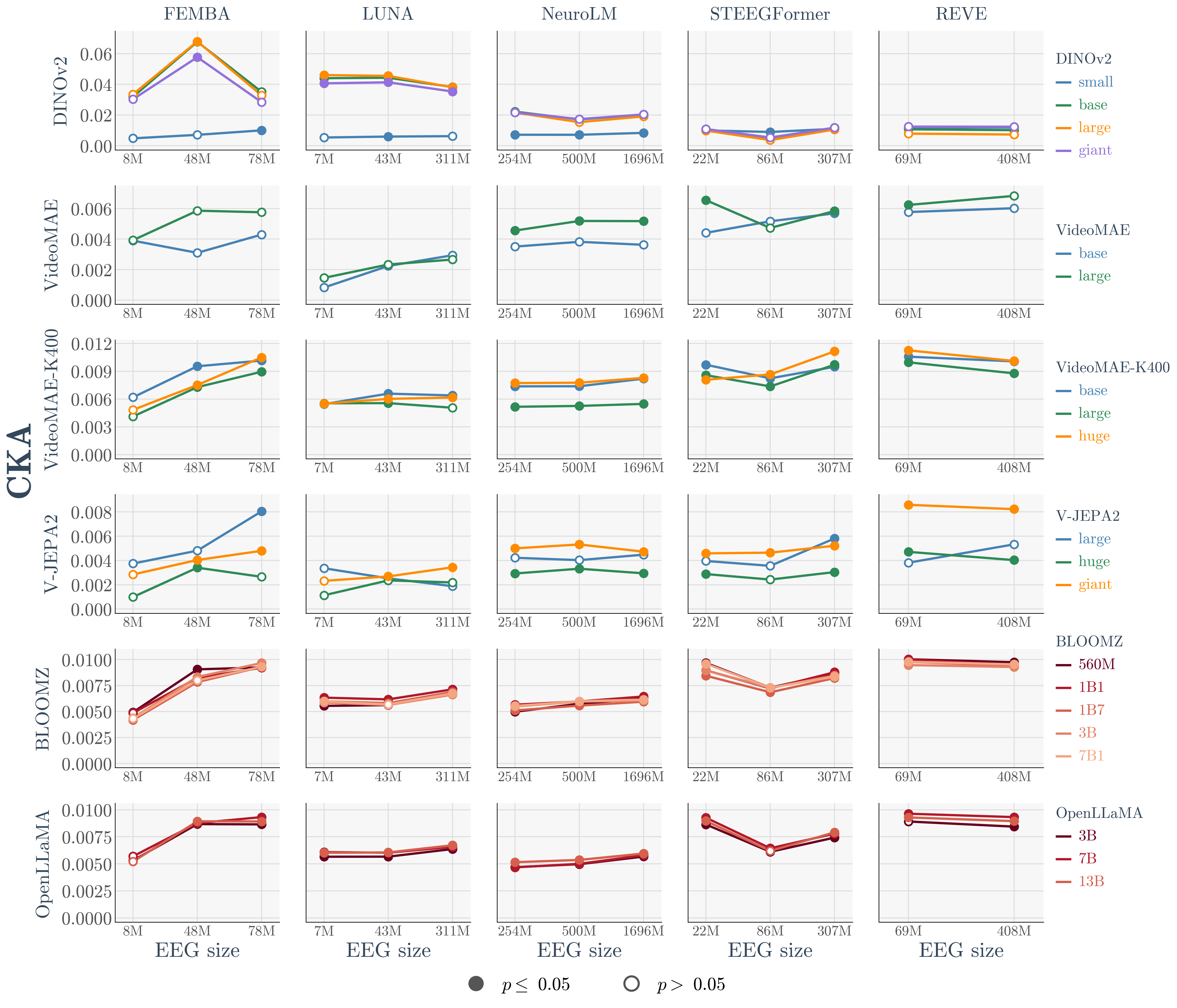}
  \caption{Each EEG model CKA alignment to every vision and language model. Many exceed the null but there is no correlation with model size.}
  \label{fig:all_vision_lang_calibrated_cka}
\end{figure}

Each EEG model is paired with each vision model on the EEG model's own tiling and scored with mKNN in Figure \ref{fig:all_vision_lang_calibrated_mknn} and with CKA in Figure \ref{fig:all_vision_lang_calibrated_cka}, taking the best score over layer pairs and calibrating against the cyclic null. Two questions are separable here, and we treat them separately: whether alignment exists at all, and whether it grows with model capacity. Although they don't always clear the null with $p > 0.05$, alignment between EEG models and models not trained on neural data therefore exists in many cases, but it doesn't improve as the EEG models grow.  

This may be explainable by the results of \citet{koepke2026platoscave}, where they argue that in a non-bijective dataset the alignment seems to be more fragile. This seems to be valid in the CineBrain dataset, as all stimuli come from the same series.

\section{Alignment of vision and language models with fMRI} 

\begin{figure}[H]
  \centering
  \includegraphics[width=\textwidth]{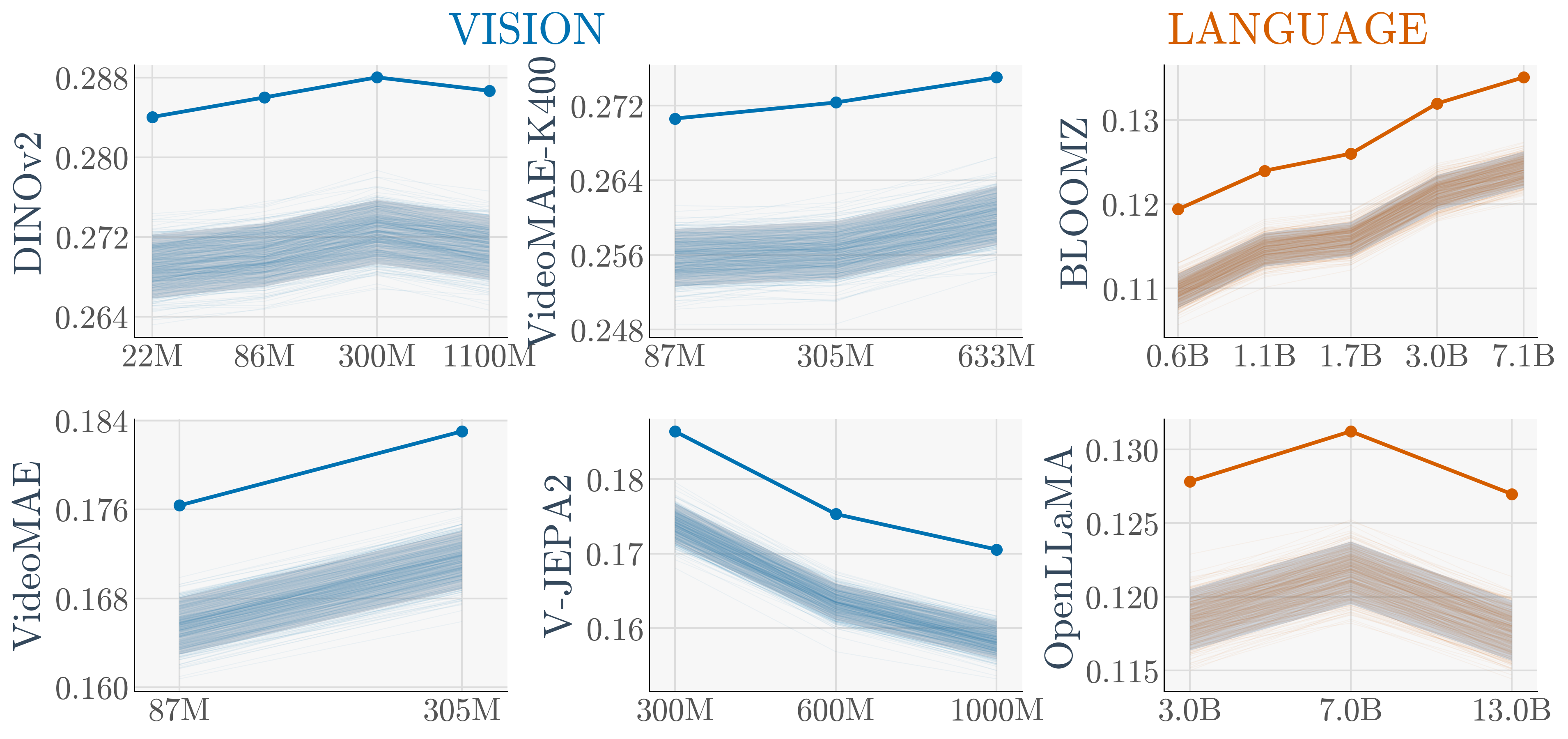}
   \vspace{-1.3em}
\caption{\textbf{mKNN alignment to brain activity against model size.} Similarity of each vision and
language model to fMRI, plotted against parameter count. Solid coloured lines show observed mKNN at the true temporal pairing. The raw score changes with size, but the null tracks it closely.}
  \label{fig:fmri_sizescaling}
\end{figure}

Figure \ref{fig:fmri_sizescaling} shows the alignment between fMRI and video and language models under mKNN. The margin above the null is considerably larger than for EEG (Fig. \ref{fig:femba_dinov2}). This is consistent with fMRI's high spatial resolution and its well-documented capacity to encode rich semantic \cite{huth2016natural} and hierarchical visual \cite{devereux2018integrated} structure. Within each model family, however, the null tracks the observed line almost exactly. As with EEG, what the larger model gains with model size is shared temporal smoothness.

%% file: refs.bib
@inproceedings{huh2024platonic,
  title     = {The Platonic Representation Hypothesis},
  author    = {Huh, Minyoung and Cheung, Brian and Wang, Tongzhou and Isola, Phillip},
  booktitle = {International Conference on Machine Learning (ICML)},
  year      = {2024},
  note      = {arXiv:2405.07987}
}

@article{groger2026aristotelian,
  title   = {Revisiting the Platonic Representation Hypothesis: An Aristotelian View},
  author  = {Gr{\"o}ger, Fabian and Wen, Shuo and Brbi{\'c}, Maria},
  journal = {arXiv preprint arXiv:2602.14486},
  year    = {2026}
}

@article{dicarlo2012how,
  author  = {DiCarlo, James J. and Zoccolan, Davide and Rust, Nicole C.},
  title   = {How Does the Brain Solve Visual Object Recognition?},
  journal = {Neuron},
  year    = {2012},
  volume  = {73},
  number  = {3},
  pages   = {415--434},
  doi     = {10.1016/j.neuron.2012.01.010},
}

@article{grillspector2004human,
  author  = {Grill-Spector, Kalanit and Malach, Rafael},
  title   = {The Human Visual Cortex},
  journal = {Annual Review of Neuroscience},
  year    = {2004},
  volume  = {27},
  pages   = {649--677},
  doi     = {10.1146/annurev.neuro.27.070203.144220},
}

@article{patterson2007where,
  author  = {Patterson, Karalyn and Nestor, Peter J. and Rogers, Timothy T.},
  title   = {Where Do You Know What You Know? {The} Representation of Semantic Knowledge in the Human Brain},
  journal = {Nature Reviews Neuroscience},
  year    = {2007},
  volume  = {8},
  number  = {12},
  pages   = {976--987},
  doi     = {10.1038/nrn2277},
}

@article{lambonralph2017neural,
  author  = {{Lambon Ralph}, Matthew A. and Jefferies, Elizabeth and Patterson, Karalyn and Rogers, Timothy T.},
  title   = {The Neural and Computational Bases of Semantic Cognition},
  journal = {Nature Reviews Neuroscience},
  year    = {2017},
  volume  = {18},
  number  = {1},
  pages   = {42--55},
  doi     = {10.1038/nrn.2016.150},
}

@article{mitchell2008predicting,
  author  = {Mitchell, Tom M. and Shinkareva, Svetlana V. and Carlson, Andrew and Chang, Kai-Min and Malave, Vicente L. and Mason, Robert A. and Just, Marcel Adam},
  title   = {Predicting Human Brain Activity Associated with the Meanings of Nouns},
  journal = {Science},
  year    = {2008},
  volume  = {320},
  number  = {5880},
  pages   = {1191--1195},
  doi     = {10.1126/science.1152876},
}

@article{huth2016natural,
  title={Natural speech reveals the semantic maps that tile human cerebral cortex},
  author={Huth, Alexander G. and de Heer, Wendy A. and Griffiths, Thomas L. and Theunissen, Fr{\'e}d{\'e}ric E. and Gallant, Jack L.},
  journal={Nature},
  volume={532},
  number={7600},
  pages={453--458},
  year={2016},
  publisher={Nature Publishing Group},
  doi={10.1038/nature17637}
}

@article{devereux2018integrated,
  title={Integrated deep visual and semantic attractor neural networks predict {fMRI} pattern-information along the ventral object processing pathway},
  author={Devereux, Barry J. and Clarke, Alex and Tyler, Lorraine K.},
  journal={Scientific Reports},
  volume={8},
  number={1},
  pages={10636},
  year={2018},
  publisher={Nature Publishing Group},
  doi={10.1038/s41598-018-28865-1}
}

@article{harris1954distributional,
  author  = {Harris, Zellig S.},
  title   = {Distributional Structure},
  journal = {Word},
  year    = {1954},
  volume  = {10},
  number  = {2-3},
  pages   = {146--162},
}

@article{yashwante2026timeseries,
  title={Time Series, Vision, and Language: Exploring the Limits of Alignment in Contrastive Representation Spaces},
  author={Yashwante, Pratham and Yu, Rose},
  journal={arXiv preprint arXiv:2602.19367},
  year={2026}
}

@article{koepke2026platoscave,
  title={Back into Plato's Cave: Examining Cross-modal Representational Convergence at Scale},
  author={Koepke, A. Sophia and Zverev, Daniil and Ginosar, Shiry and Efros, Alexei A.},
  journal={arXiv preprint arXiv:2604.18572},
  year={2026}
}

@article{gao2025cinebrain,
  title   = {CineBrain: A Large-Scale Multi-Modal Brain Dataset During Naturalistic Audiovisual Narrative Processing},
  author  = {Gao, Jianxiong and Liu, Yichang and Yang, Baofeng and Feng, Jianfeng and Fu, Yanwei},
  journal = {arXiv preprint arXiv:2503.06940},
  year    = {2025}
}

@inproceedings{kornblith2019cka,
  title     = {Similarity of Neural Network Representations Revisited},
  author    = {Kornblith, Simon and Norouzi, Mohammad and Lee, Honglak and Hinton, Geoffrey},
  booktitle = {International Conference on Machine Learning (ICML)},
  year      = {2019}
}

@article{tegon2025femba,
  title   = {{FEMBA}: Efficient and Scalable {EEG} Analysis with a Bidirectional Mamba Foundation Model},
  author  = {Tegon, Anna and Ingolfsson, Thorir Mar and Wang, Xiaying and Benini, Luca and Li, Yawei},
  journal = {arXiv preprint arXiv:2502.06438},
  year    = {2025}
}

@inproceedings{doner2025luna,
  title     = {{LUNA}: Efficient and Topology-Agnostic Foundation Model for {EEG} Signal Analysis},
  author    = {D{\"o}ner, Berkay and Ingolfsson, Thorir Mar and Benini, Luca and Li, Yawei},
  booktitle = {Advances in Neural Information Processing Systems (NeurIPS)},
  year      = {2025},
  note      = {arXiv:2510.22257}
}

@article{Bandt2002,
  title = {Permutation Entropy: A Natural Complexity Measure for Time Series},
  author = {Bandt, Christoph and Pompe, Bernd},
  journal = {Phys. Rev. Lett.},
  volume = {88},
  issue = {17},
  pages = {174102},
  numpages = {4},
  year = {2002},
  month = {Apr},
  publisher = {American Physical Society},
  doi = {10.1103/PhysRevLett.88.174102},
  url = {https://link.aps.org/doi/10.1103/PhysRevLett.88.174102}
}

@article{elouahidi2025reve,
  title   = {{REVE}: A Foundation Model for {EEG} -- Adapting to Any Setup with Large-Scale Pretraining on 25,000 Subjects},
  author  = {El Ouahidi, Yassine and Lys, Jonathan and Th{\"o}lke, Philipp and Farrugia, Nicolas and Pasdeloup, Bastien and Gripon, Vincent and Jerbi, Karim and Lioi, Giulia},
  journal = {arXiv preprint arXiv:2510.21585},
  year    = {2025}
}

@inproceedings{jiang2025neurolm,
  title     = {{NeuroLM}: A Universal Multi-task Foundation Model for Bridging the Gap between Language and {EEG} Signals},
  author    = {Jiang, Wei-Bang and Wang, Yansen and Lu, Bao-Liang and Li, Dongsheng},
  booktitle = {International Conference on Learning Representations (ICLR)},
  year      = {2025},
  note      = {arXiv:2409.00101}
}

@article{oquab2023dinov2,
  title   = {{DINOv2}: Learning Robust Visual Features without Supervision},
  author  = {Oquab, Maxime and Darcet, Timoth{\'e}e and Moutakanni, Th{\'e}o and Vo, Huy and Szafraniec, Marc and Khalidov, Vasil and others},
  journal = {arXiv preprint arXiv:2304.07193},
  year    = {2023}
}

@inproceedings{tong2022videomae,
  title     = {{VideoMAE}: Masked Autoencoders are Data-Efficient Learners for Self-Supervised Video Pre-Training},
  author    = {Tong, Zhan and Song, Yibing and Wang, Jue and Wang, Limin},
  booktitle = {Advances in Neural Information Processing Systems (NeurIPS)},
  year      = {2022}
}

@article{assran2025vjepa2,
  title   = {{V-JEPA} 2: Self-Supervised Video Models Enable Understanding, Prediction and Planning},
  author  = {Assran, Mido and Bardes, Adrien and Fan, David and Garrido, Quentin and Howes, Russell and others},
  journal = {arXiv preprint arXiv:2506.09985},
  year    = {2025}
}

@misc{geng2023openllama,
  title  = {{OpenLLaMA}: An Open Reproduction of {LLaMA}},
  author = {Geng, Xinyang and Liu, Hao},
  year   = {2023},
  howpublished = {\url{https://github.com/openlm-research/open_llama}}
}

@article{sitt2014large,
  title   = {Large scale screening of neural signatures of consciousness in patients in a vegetative or minimally conscious state},
  author  = {Sitt, Jacobo D and King, Jean-R{\'e}mi and El Karoui, Imen and Rohaut, Benjamin and Faugeras, Frederic and Gramfort, Alexandre and Cohen, Laurent and Sigman, Mariano and Dehaene, Stanislas and Naccache, Lionel},
  journal = {Brain},
  volume  = {137},
  number  = {8},
  pages   = {2258--2270},
  year    = {2014}
}

@article{engemann2018robust,
  title   = {Robust EEG-based cross-site and cross-protocol classification of states of consciousness},
  author  = {Engemann, Denis A and Raimondo, Federico and King, Jean-R{\'e}mi and Rohaut, Benjamin and Louppe, Gilles and Faugeras, Frederic and Annen, Jitka and Cassol, Helena and Gosseries, Olivia and Fernandez-Slezak, Diego and Laureys, Steven and Naccache, Lionel and Dehaene, Stanislas and Sitt, Jacobo D},
  journal = {Brain},
  volume  = {141},
  number  = {11},
  pages   = {3179--3192},
  year    = {2018}
}

@inproceedings{yang2026steegformer,
  title     = {Are {EEG} Foundation Models Worth It? Comparative Evaluation with Traditional Decoders in Diverse {BCI} Tasks},
  author    = {Yang, Liuyin and Sun, Qiang and Li, Ang and Van Hulle, Marc M.},
  booktitle = {International Conference on Learning Representations (ICLR)},
  year      = {2026},
  url       = {https://openreview.net/forum?id=5Xwm8e6vbh}
}

@inproceedings{borrell2026universe,
  title     = {The Platonic Universe: Do Foundation Models See the Same Sky?},
  author    = {Borrell, Trinidad and Dillmann, Steven and Duraphe, Kshitij and Eris, Furkan and Khederlarian, Ashod and Kumar, Aman and Marraffini, Giovanni and Smith, Michael J. and Sourav, Shashwat and Di Tella, Rocco and Wu, John F.},
  booktitle = {Mechanistic Interpretability Workshop, International Conference on Machine Learning (ICML)},
  year      = {2026}
}

@article{kaplan2020scaling,
  title={Scaling Laws for Neural Language Models},
  author={Kaplan, Jared and McCandlish, Sam and Henighan, Tom and Brown, Tom B.
    and Chess, Benjamin and Child, Rewon and Gray, Scott and Radford, Alec
    and Wu, Jeffrey and Amodei, Dario},
  journal={arXiv preprint arXiv:2001.08361},
  year={2020}
}

@inproceedings{hoffmann2022chinchilla,
  title={Training Compute-Optimal Large Language Models},
  author={Hoffmann, Jordan and Borgeaud, Sebastian and Mensch, Arthur and others},
  booktitle={Advances in Neural Information Processing Systems},
  year={2022}
}

@inproceedings{bloomz,
  title     = {Crosslingual Generalization through Multitask Finetuning},
  author    = {Muennighoff, Niklas and Wang, Thomas and Sutawika, Lintang and
               Roberts, Adam and Biderman, Stella and Le Scao, Teven and
               Bari, M Saiful and Shen, Sheng and Yong, Zheng Xin and
               Schoelkopf, Hailey and Tang, Xiangru and Radev, Dragomir and
               Aji, Alham Fikri and Almubarak, Khalid and Albanie, Samuel and
               Alyafeai, Zaid and Webson, Albert and Raff, Edward and Raffel, Colin},
  booktitle = {Proceedings of the 61st Annual Meeting of the Association for
               Computational Linguistics (Volume 1: Long Papers)},
  pages     = {15991--16111},
  year      = {2023},
  address   = {Toronto, Canada},
  publisher = {Association for Computational Linguistics},
  doi       = {10.18653/v1/2023.acl-long.891},
  url       = {https://aclanthology.org/2023.acl-long.891/}
}
